\documentclass[10pt,twocolumn,letterpaper]{article}

 \usepackage[pagenumbers]{iccv} 

\usepackage{amsfonts, amssymb}
\usepackage{amsmath} 
\DeclareMathOperator*{\argmax}{arg\,max}

\definecolor{iccvblue}{rgb}{0.21,0.49,0.74}
\usepackage[pagebackref,breaklinks,colorlinks,allcolors=iccvblue]{hyperref}
\usepackage{algorithm}
\usepackage{tikz}
\usepackage{algpseudocode}
\usepackage{amsmath}
\usepackage{amsfonts}
\usepackage{amssymb}
\usepackage{marvosym}
\usepackage{booktabs} 
\usepackage{graphicx} 
\usepackage{multirow}

\def\paperID{13619} 
\def\confName{ICCV}
\def\confYear{2025}
\DeclareMathOperator*{\argmin}{arg\,min}
\title{Integrating Task-Specific and Universal Adapters for \\ Pre-Trained Model-based Class-Incremental Learning}

\author{
Yan Wang,
Da-Wei Zhou\textsuperscript{(\Letter)},
Han-Jia Ye
\and
School of Artificial Intelligence, Nanjing University\\
National Key Laboratory for Novel Software Technology, Nanjing University\\
{\tt\small {\{wangy,zhoudw,yehj\}@lamda.nju.edu.cn}}
}

\begin{document}
\maketitle
\footnotetext[2]{Correspondence to: Da-Wei Zhou (zhoudw@lamda.nju.edu.cn)}

\begin{abstract}
Class-Incremental Learning (CIL) requires a learning system to continually learn new classes without forgetting. Existing pre-trained model-based CIL
methods often freeze the pre-trained network and adapt to incremental tasks using additional lightweight modules such as adapters. However, incorrect module selection during inference hurts performance, and task-specific modules often overlook shared general knowledge, leading to errors on distinguishing between similar classes across tasks. To address the aforementioned challenges, we propose integrating Task-Specific and Universal Adapters (TUNA) in this paper. Specifically, we train task-specific adapters to capture the most crucial features relevant to their respective tasks and introduce an entropy-based selection mechanism to choose the most suitable adapter. Furthermore, we leverage an adapter fusion strategy to construct a universal adapter, which encodes the most discriminative features shared across tasks. We combine task-specific and universal adapter predictions to harness both specialized and general knowledge during inference. Extensive experiments on various benchmark datasets demonstrate the state-of-the-art performance of our approach. Code is available at \url{https://github.com/LAMDA-CL/ICCV2025-TUNA}
\end{abstract}.    

\section{Introduction}
The advent of deep learning leads to the remarkable performance of deep neural networks in practical applications~\cite{deng2009imagenet,chen2022learning,ye2024bridging,gansemi}. However, in real-world scenarios, data often arrive in a continuous stream, necessitating a learning system capable of progressively assimilating knowledge of emerging classes, a process known as class-incremental learning (CIL)~\cite{Rebuffi_2017_CVPR}. CIL faces a formidable challenge: the process of acquiring new classes often results in the erosion of previously learned knowledge, precipitating a phenomenon known as catastrophic forgetting of established features~\cite{french1999catastrophic}. Correspondingly, recent breakthroughs in pre-training~\cite{han2021pre} have prompted the research community to leverage pre-trained models (PTMs) as a means to mitigate the issue of forgetting~\cite{wang2022learning,wang2022dualprompt,Tan_2024_CVPR}. Leveraging vast datasets and considerable computing resources, PTMs naturally produce features with strong generalization capabilities. As a result, the development of a robust CIL methodology that harnesses the power of PTMs while mitigating catastrophic forgetting has attracted considerable attention from the research community~\cite{wang2022s,zhou2024revisiting,zhou2024expandable,zhou2025learning}.

Owing to the remarkable generalization properties of PTMs, existing approaches frequently involve freezing the pre-trained weights and adapting to incremental tasks through the integration of lightweight modules ~\cite{jia2022visual,lian2022scaling,chen2022adaptformer}. Many of these methods rely on visual prompt tuning~\cite{wang2022dualprompt, Smith_2023_CVPR}. During training, they learn task-specific prompt parameters and a set of keys. These keys are later used for task selection through query-key matching during inference. However, these methods suffer from two drawbacks. First, continual learning requires models to dynamically adapt to a sequence of tasks while maintaining stability on previous ones. Current prompt-based methods rely on accurate retrieval of task-specific prompts during inference. However, incorrect key matching, especially in scenarios with task ambiguity or distribution shifts, can lead to the selection of irrelevant prompts, degrading performance. Second, these methods predominantly concentrate on the acquisition of task-specific knowledge and ignore the general knowledge shared between different tasks. Thus, they tend to make mistakes when distinguishing between highly similar classes across different tasks.

To overcome the challenges outlined above, we introduce integrating Task-Specific and Universal Adapters (TUNA) in this paper, which explicitly disentangles continual learning into two complementary components: (1) specialized adapters that extract task-discriminative features, and (2) a universal adapter that consolidates cross-task shared knowledge through fusion. This decomposition not only mitigates task interference but also enables more robust generalization to semantically overlapping classes.

First, we use orthogonal loss to train task-specific adapters. To enhance the accuracy of module selection, we introduce an entropy-based adapter selection strategy that routes inputs to the most relevant task-specific adapter based on prediction uncertainty, eliminating reliance on brittle key-query matching. Second, for knowledge consolidation, we leverage an adapter fusion technique that merges task-specific adapters into a universal adapter, preserving shared features while minimizing redundancy. During inference, our framework leverages both task-specific and universal adapters in a coordinated manner. Our comprehensive experiments validate that the proposed method achieves state-of-the-art results across benchmark datasets, demonstrating notable improvements on challenging datasets such as ImageNet-A and ObjectNet.

\section{Related Work}
\noindent\textbf{Class-Incremental Learning (CIL):} necessitates a learning system capable of continuously assimilating new class information while preserving previously acquired knowledge without forgetting~\cite{goswami2023fecam,zheng2025task,zheng2024multi,li2025addressing,liu2023online,luo2023class}. This paradigm can be broadly categorized into several categories. Data rehearsal-based methods~\cite{aljundi2019gradient,zhao2021memory,rolnick2019experience,chaudhry2018riemannian,chaudhry2021using} involve carefully selecting and reintroducing exemplars from earlier classes during the acquisition of new classes. Knowledge distillation-based methods~\cite{li2017learning,hou2018lifelong,kang2022class,douillard2020podnet,simon2021learning} try to establish
 a mapping between the model from previous stages and the current model through the process of knowledge distillation~\cite{hinton2015distilling}. These mappings, represented as logits or feature representations, assit the incremental model in retaining essential characteristics from earlier phases during updating. Model rectification-based methods ~\cite{wu2019large,yu2020semantic,zhao2020maintaining,pernici2021class,ahn2021ss} seek to rectify the inductive bias inherent in incremental models, ensuring unbiased predictions during the updating process. Moreover, parameter regularization-based methods~\cite{zenke2017continual,aljundi2018memory,lee2017overcoming,lee2020continual} impose regularization penalties on the drift of crucial parameters throughout model adaptation, thereby safeguarding earlier knowledge. Expandable networks have recently shown strong performance in incremental learning~\cite{douillard2022dytox,hu2023dense,wang2022foster,yan2021dynamically}. These methods preserve the original backbone and initialize a new one for each task, combining their outputs into a large feature map and training a classifier with exemplars for calibration.

  \noindent\textbf{Pre-Trained Model-Based CIL:} is now a hot topic in today's CIL field. Most pre-trained model-based CIL methods utilize the parameter-efficient fine-tuning mechanism to
 adapt the model efficiently while keeping the pre-trained model frozen. L2P~\cite{wang2022learning} leverages a pre-trained model and dynamically learns a prompt pool to guide the model in addressing specific tasks. DualPrompt~\cite{wang2022dualprompt} introduces a novel approach by learning two mutually independent prompt spaces: the general prompt and the expert prompt, which encode task-invariant and task-specific knowledge, respectively. CODA-Prompt~\cite{Smith_2023_CVPR} presents a decomposition-based, attention-driven continual learning prompting method, offering a significantly larger learning capacity compared to existing prompt-based techniques. Instead of directly optimizing prompt parameters, DAP~\cite{jung2023generating} designs prompt generators to generate instance-specific information in prompts. SLCA~\cite{zhang2023slca} employs distinct learning rates for the backbone and classifier, it also models class-wise feature distributions~\cite{zhu2021prototype} and replays them to calibrate the classifier. APER~\cite{zhou2024revisiting} proposes constructing the classifier by merging embeddings from both the pre-trained model and the adapted downstream model. EASE~\cite{zhou2024expandable} innovatively concatenates feature representations from multiple task-specific backbones, further enhancing model capabilities.
  Furthermore, RanPAC~\cite{McDonnell_2023_NeurIPS} introduces a random projection approach that constructs robust high-dimensional randomized features, proving effective for continual learning tasks.
\section{Preliminaries}
 In this section, we introduce the background of class-incremental learning and corresponding baselines.
\subsection{Class-Incremental Learning}
CIL focuses on continuously learning from evolving data streams that introduce new classes, while preserving the knowledge of previously encountered classes to construct a unified classifier~\cite{Rebuffi_2017_CVPR}. Consider a series of $T$ training stages, represented as $\{\mathcal{D}^1,\mathcal{D}^2,\cdots,\mathcal{D}^T\}$,where $\mathcal{D}^t=\{(\mathbf{x}_{i}^{t}, y_{i}^{t})\}_{i=1}^{n_t}$ denotes the $t$-th incremental stage containing $n_t$ instances.
Correspondingly, the testing set is denoted as $\{\mathcal{D}_t^1,\mathcal{D}_t^2,\cdots,\mathcal{D}_t^T\}$. Each training instance $\mathbf{x}_{i}^{t} \in \mathcal{D}^t$ is associated with a class label $y_{i}^{t} \in Y_t$. Here, $Y_t$ defines the set of labels for training task $t$, and it is guaranteed that $Y_t \cap Y_{t^{\prime}}=\varnothing$ for any $t\neq t^\prime$. In this paper, we follow the \textbf{exemplar-free setting} in~\cite{zhou2024expandable}, which means that no historical exemplars from previous classes are used. Consequently, the model only has access to data from $\mathcal{D}^t$ for training during the $t$-th stage. The model's performance is evaluated across all previously encountered classes, denoted as $\mathcal{Y}_t = Y_1 \cup \cdots \cup Y_t$, after each incremental learning task. Specifically, our objective is to learn a model $f(\mathbf{x}): X \rightarrow \mathcal{Y}_t$ that minimizes empirical risk across all test classes:
\begin{equation} \label{eq:cilrisk} 
f^*=\underset{f\in\mathcal{H}}{\operatorname*{argmin}} \ \mathbb{E}_{(\mathbf{x},y)\sim\mathcal{D}_t^1\cup\cdots\mathcal{D}_t^t}\mathbb{I}\left(y\neq f(\mathbf{x})\right),
\end{equation}
where $\mathcal{H}$ is the hypothesis space and $\mathbb{I}(\cdot)$ denotes the indicator function, $\mathcal{D}_t^b$ refers to the testing set for task $b$. An effective CIL model satisfying Eq.~\ref{eq:cilrisk} demonstrates strong discriminative abilities across all classes. It strikes a balance between acquiring knowledge of new classes and preserving information from previously learned ones.

In line with typical PTM-based CIL works~\cite{wang2022learning,wang2022dualprompt,zhou2024expandable}, we assume that a pre-trained Vision Transformer (ViT)~\cite{dosovitskiy2020vit} is available as the initialization for $f(\mathbf{x})$. To facilitate understanding, we decompose the PTM into two components: $f(\mathbf{x})=W^{\top}\phi(\mathbf{x})$, where $\phi(\cdot):\mathbb{R}^{D} \rightarrow \mathbb{R}^{d}$ is the feature extractor and $W\in\mathbb{R}^{d\times |\mathcal{Y}_{t}|}$ is the classifier. We denote the classifier for class $k$ as $\mathbf{w}_k$, so that $W=[\mathbf{w}_1,\mathbf{w}_2,\cdots,\mathbf{w}_{|\mathcal{Y}_{t}|}]$. 

\subsection{Baselines in PTM-Based CIL}
In the era of PTMs, many methods seek to modify the PTM slightly to maintain the pre-trained knowledge~\cite{wang2022learning,wang2022dualprompt,Smith_2023_CVPR,wang2023hierarchical}. These methods usually involve freezing the pre-trained weights and training  additional modules like  prompt pool to incorporate task-specific information. A representative example is L2P~\cite{wang2022learning}, which proposes a key-query matching strategy. Specifically, every prompt $P_i \in \mathbb{R}^{L \times d}$, with  $L$ denoting the prompt length, is associated with a learnable key vector $\mathbf{k_i}\in\mathbb{R}^d$. The prompt pool is defined as $\mathbf{P}=\{(\mathbf{k}_1, P_1),(\mathbf{k}_2, P_2),\cdots,(\mathbf{k}_Q, P_Q)\}$, where $Q$ is the pool size. The optimization objective is formulated as:
\begin{equation} \label{eq:l2p} 
\min_{W, \mathbf{P}} \sum_{(\mathbf{x}, y) \in D^t} \mathcal{L}_{ce} \left( W^\top \bar{\phi} \left( \mathbf{x}; \mathbf{P} \right), y \right) + \mathcal{L}_{reg}(\mathbf{P}),
\end{equation}
where $\bar{\phi}\left(\cdot\right)$ represents the frozen pre-trained backbone parameters, $\mathcal{L}_{ce}$ corresponds to the cross-entropy loss, and $\mathcal{L}_{reg}$ serves as the prompt selection regularization term. During inference, the most appropriate prompts are selected by identifying the top-$N$ keys:
\begin{equation} \label{eq:l2pselect}
	\mathbf{K}=\underset{\left\{t_i\right\}_{i=1}^N \subseteq[1, Q]}{\operatorname{argmin}} \quad \sum_{i=1}^N d\left(\phi({\mathbf{x}}), \mathbf{k}_{t_i}\right) \,,
\end{equation}
\noindent where $\left\{t_i\right\}_{i=1}^N$ is the selected index set, and $\mathbf{K}$ is the selected top-$N$ keys, $d(\cdot, \cdot)$ denotes the cosine distance. 

This approach has two main limitations. First, it demands precise selection of the most appropriate lightweight modules during inference, as guided by Eq.~\ref{eq:l2pselect}. However, the key-query matching process is fragile, making it prone to selecting unsuitable modules, which in turn leads to performance degradation. Second, it primarily focuses on task-specific knowledge while neglecting shared general knowledge between tasks. For instance, if the model learns to classify dogs and cats in different tasks, it may confuse similar-looking classes like a fluffy dog or a cat with a long snout due to its narrow focus on task-specific features. Consequently, it tends to make mistakes when distinguishing highly similar classes across tasks.
\section{Methodology}
To address the aforementioned challenges, we introduce TUNA in this paper. First, we train task-specific adapters and use an entropy-based mechanism to select the best adapter for each input. Second, we fuse these adapters into a universal adapter to retain shared knowledge across tasks. During inference, we employ a dual-adapter strategy that simultaneously leverages both the selected task-specific adapter and the universal adapter to boost the accuracy.
\begin{figure*}[t]
	\begin{center}
		\includegraphics[width=2.0\columnwidth]{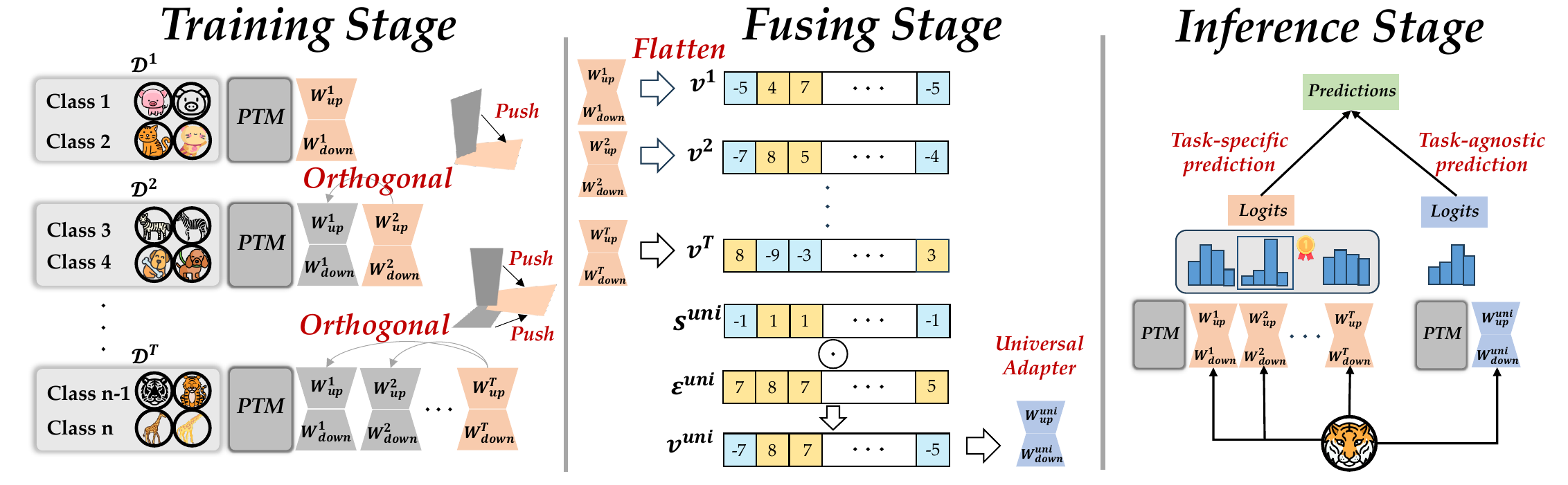}
	\end{center}
	\caption{\small  Illustration of TUNA. {\bf Left}: The training protocol of TUNA. We use orthogonal loss to train task-specific adapters.
    {\bf Middle}: The fusing process. We construct an aggregated sign vector and a magnitude vector, which are combined to form the universal task vector.
	{\bf Right}: During the inference phase, we select the most appropriate task-specific adapter based on entropy, and then combine the outputs from both the task-specific and universal adapters.
	} 
    \label{figure:method}
\end{figure*} 
\subsection{Learning Orthogonal Task-Specific Adapters} 
\noindent In this paper, we follow~\cite{zhou2024revisiting} to use  adapter~\cite{chen2022adaptformer}  to efficiently adapt the PTM to downstream tasks. An adapter is a bottleneck structure that can be incorporated into a pre-trained vision-transformer to facilitate transfer learning. Suppose we have $L$ transformer blocks in the pre-trained vision-transformer, each with a self-attention module and an MLP layer. We can insert an adapter into each block's MLP via residual connections. An adapter comprises a down-projection layer $W_{down} \in \mathbb{R}^{d \times r}$, a non-linear activation function ReLU and an up-projection layer $W_{up} \in \mathbb{R}^{r \times d}$. The output formula of the MLP layer is formulated as follows:
\begin{align} \label{eq:adapter-define}
	\mathbf{x}_o = \text{MLP}(\mathbf{x}_i)+\text{ReLU}(\mathbf{x}_i W_{down})W_{up} ,
\end{align}
where $\mathbf{x}_i$ and $\mathbf{x}_o$ are the input and output of the MLP, respectively. Eq.~\ref{eq:adapter-define} illustrates how to inject task-specific information by adding residual connections of adapters to the original outputs. For a specific task $i$, we define the set of adapters across all transformer blocks as $\mathcal{A}_i$, which represents task-specific adapters. Furthermore, we denote the output embedding of a given $\mathcal{A}_i$ combined with the PTM as $\phi(\mathbf x;\mathcal{A}_i)$, the corresponding prediction as $ f(\mathbf x;\mathcal{A}_i)=W^\top{\phi}\left(\mathbf x;\mathcal{A}_i\right)$. During the learning process of task $t$, we initialize a new adapter $\mathcal{A}_t$, which is composed of $W_{down}^t$ and $W_{up}^t$, and then freeze the weights of the PTM, focus solely on optimizing the task-specific adapters and the corresponding classifier:
\begin{align} \label{eq:adapter-opt}
    \mathcal{L}_{cls} = -\frac{1}{n_t} \sum_{(\mathbf{x}, y) \in D^t} \log \frac{\exp(\mathbf{w}_y^\top\phi(\mathbf x;\mathcal{A}_t))}{\sum_{i=1}^{|\mathcal{Y}_{t}|} \exp(\mathbf{w}_i^\top\phi(\mathbf x;\mathcal{A}_t))},
\end{align}
where $n_t$ denotes the number of instances in $t$-th incremental stage.
After the first task, we utilize an orthogonal loss function to ensure that the trainable weights remain orthogonal to those learned from previous tasks:
\begin{align} \label{eq:adapter-orth}
    \mathcal{L}_{orth} =\sum_{i=1}^{t-1}\left\| W_{up}^{t} \cdot  {W_{up}^{i}}^\top \right\|_{1},
\end{align}
where $\left\| \cdot \right\|_{1}$ represents the $L_1$ norm. The up-projection weights in the adapter module plays a key role in projecting intermediate features into a higher-dimensional space, which is essential for encoding task-specific information. By imposing orthogonality constraints on the up-projection weights, we ensure that the current adapter learns \emph{unique} and non-redundant features, effectively differentiating it from previously learned adapters. The overall optimization target is formulated as:
\begin{align} \label{eq:adapter-overall}
    \mathcal{L} = \mathcal{L}_{cls} +
    \lambda\mathcal{L}_{orth},
\end{align}
where $\lambda$ is a scalar to weight the loss. After training $T$ tasks by optimizing Eq.~\ref{eq:adapter-overall}, we get a list of $T$ adapters: $\{\mathcal{A}_1,\mathcal{A}_2,\cdots,\mathcal{A}_T \}$. These adapters effectively capture the most salient features for their respective tasks.

\noindent\textbf{Effect of task-specific adapters:}
Figure~\ref{figure:method} (Left) shows the training protocol. We independently train and optimize adapter modules for each incremental task, ensuring each module extracts maximally discriminative task-specific features. This framework is general and can be seamlessly integrated with various parameter-efficient fine-tuning techniques like LoRA~\cite{hu2022lora} and VPT~\cite{jia2022visual}. Additionally, the lightweight architecture of adapters requires significantly fewer trainable parameters than full-model fine-tuning.

\subsection{Multi-Stage Adapter Fusion} \label{sec:fusion}
\noindent After training on $t$ tasks, we obtain a set of task-specific adapters $\{\mathcal{A}_1,\mathcal{A}_2,\cdots,\mathcal{A}_t \}$. These adapters 
are derived by optimizing the same PTM via Eq.~\ref{eq:adapter-overall}, ensuring that each adapter is discriminative for its respective task and functions as a ‘task expert.’ For example, if the first task involves classifying ‘tigers,’
 the first adapter will focus on features like furpatterns and stripes. If the next task contains ‘birds,’ the adapter will emphasize characteristics such as beaks and feathers. Thus, each adapter is typically limited to task-specific knowledge and struggles to differentiate between similar classes across tasks. In a simplified scenario where the task identity is known, we could directly use the corresponding expert adapter for prediction. However, in class-incremental learning, where task identity is not available, it is necessary to create a unified embedding space that accommodates all tasks. Drawing on insights from model merging techniques~\cite{matena2022merging,yang2023adamerging,huang_emrmerging}, we want to integrate these task-specific adapters into a \emph{universal} adapter that can capture the high-level features shared across all tasks.

To achieve this, we begin by flattening the weights of the task-specific adapters into vectors: $ \mathbf{v}^i = \text{Flatten}(\mathcal{A}_i)$, resulting in a collection of task-specific vectors, denoted as $\{\mathbf{v}^1,\mathbf{v}^2,\cdots,\mathbf{v}^t \}$. Next, we construct the universal sign vector  by determining the dominant sign for each parameter across all task-specific vectors. This is done by taking the sign of the sum of the corresponding parameters: 
\begin{align}
     \mathbf{s}^{\text{uni}} = \text{sgn}\left(\sum_{i=1}^t \mathbf{v}^i\right),
\end{align}

\noindent where $\text{sgn}(\cdot)$ denotes the sign function. For each parameter, we then identify the maximum absolute value among all task vectors that maintain the consensus sign direction, forming the magnitude vector. Specifically, the $j$-th dimension of the magnitude vector $\mathbf{\epsilon}_{j}^{\text{uni}}$ is calculated as:
\begin{align}
     \mathbf{\epsilon}_{j}^{\text{uni}} = 
\begin{cases} 
abs\left(\max(v_j^1,\cdots,v_j^t)\right) & \text{if } s_j^{\text{uni}} > 0 \\
abs\left( \min(v_j^1,\cdots,v_j^t)\right) & \text{if } s_j^{\text{uni}} < 0 
\end{cases},
\end{align}
where $s_j^{\text{uni}}$ denotes the $j$-th dimension of the sign vector. Then the universal task vector is generated through Hadamard (element-wise) multiplication: 
  \begin{align}
     \mathbf{v}^{\text{uni}} = \mathbf{\epsilon}^{\text{uni}} \odot \mathbf{s}^{\text{uni}}.
\end{align}
  Finally, we reshape $\mathbf{v}^{\text{uni}}$ to match the original dimensions of the adapter, yielding the universal adapter $\mathcal{A}_{\text{uni}}$.

\noindent\textbf{Effect of the universal adapter:}
Figure~\ref{figure:method} (Middle) illustrates the fusion process, which employs two principled operations: sign summation and max-absolute-value selection. The sign summation operates as a voting system that maintains dominant feature orientations across tasks. Concurrently, the max-absolute-value selection with sign consistency suppresses noisy minor activations while preserving task-specific feature magnitudes without attenuation. This operation is theoretically grounded in max-out networks ~\cite{goodfellow2013maxout} and has been shown to preserve discriminative features.

Thorough the sign and max operation, the resulting universal adapter captures high-level features common to all tasks, which may not be fully represented by individual task-specific adapter. By using the universal adapter, we can effectively leverage shared knowledge and ensure the model is better equipped to handle all encountered tasks.

\subsection{Adapter Selection via Prediction Uncertainty}
\begin{figure}[t]
	\centering
	\begin{subfigure}{0.49\linewidth}
		\includegraphics[width=1\columnwidth]{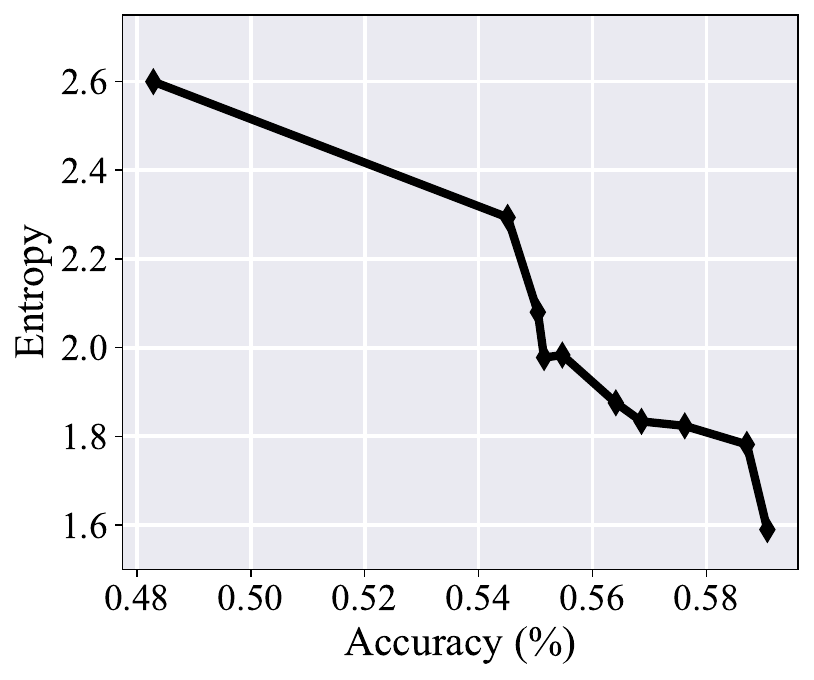}
		\caption{ImageNet-A B0 inc20}
		\label{fig:1}
	\end{subfigure}
	\hfill
	\begin{subfigure}{0.49\linewidth}
		\includegraphics[width=1\linewidth]{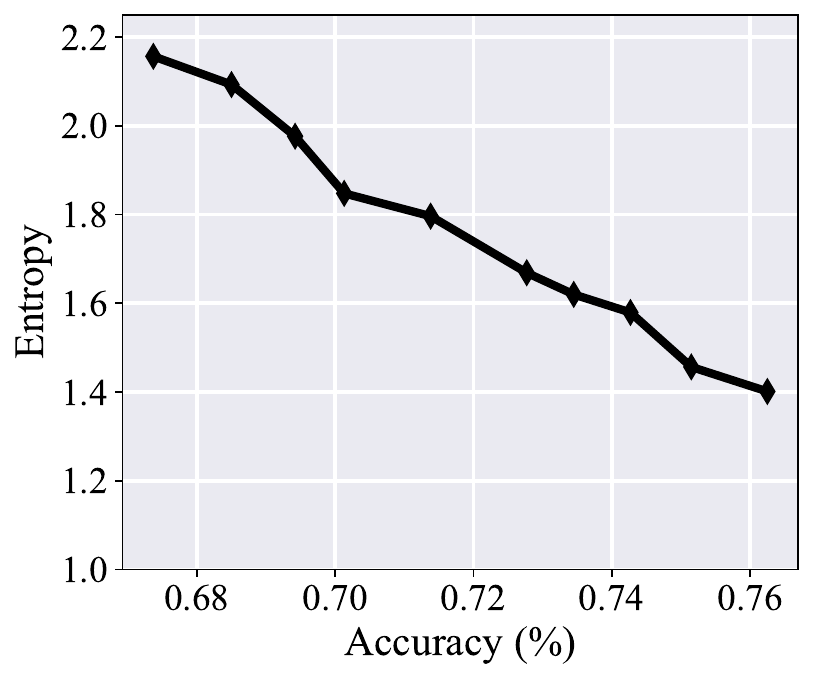}
		\caption{ImageNet-R B0 inc20}
		\label{figure:2}
	\end{subfigure}
	\caption{\small  Relationship between accuracy and entropy.}
\end{figure}

Suppose the model has progressively learned $t$ tasks and is now required to classify a test image, which may belong to any of the previously learned $t$ tasks. The primary challenge lies in selecting the most suitable task-specific adapter for this prediction.
For a sample $\mathbf{x}$, the predictions of PTM combined with different task-specific adapters 
are denoted as $f(\mathbf{x};\mathcal{A}_1),f(\mathbf{x};\mathcal{A}_2),\cdots, f(\mathbf{x};\mathcal{A}_t)$. Previous research have observed that minimizing entropy on test samples during optimization enables the pre-trained model to effectively adjust to previously unseen test data distributions. Nevertheless, it is still unclear whether entropy minimization can reliably function as a proxy objective for identifying the optimal task-specific adapter. To investigate this, we conduct a pilot study. Specifically, we choose Imagenet-A~\cite{hendrycks2021unnatural} and Imagenet-R~\cite{hendrycks2021many} as the datasets and split them into 10 tasks. We assess the model's performance when combined with each of the 10 different task-specific adapters:
$f(\mathbf{x};\mathcal{A}_1),f(\mathbf{x};\mathcal{A}_2),\cdots, f(\mathbf{x};\mathcal{A}_{10})$ respectively. We compute the corresponding entropy and prediction accuracy. As illustrated in Figure~\ref{fig:1} and Figure~\ref{figure:2}, lower entropy is associated with higher prediction accuracy. In other words, the greater the model’s confidence in its predictions, the more accurate it tends to be. Consequently, we conclude that entropy minimization effectively acts as a robust proxy objective for identifying the optimal task-specific adapter. When we have $t$ task-specific adapters, we can choose the most suitable adapter $\mathcal{A}^*$ according to the following formula:
\begin{align} \label{eq:min_entropy}
    \mathcal{A}^* = \argmin_{\mathcal{A}_i \in \{\mathcal{A}_1, \mathcal{A}_2, \dots, \mathcal{A}_t\}} \left( -\sum_{c=1}^{\mathcal{Y}_t} f_c(\mathbf{x}; \mathcal{A}_i) \log f_c(\mathbf{x}; \mathcal{A}_i) \right),
\end{align}
where $f_c(\mathbf{x}; \mathcal{A}_i)$ denotes the predicted probability of class $c$ for input $\mathbf{x}$ using adapter $\mathcal{A}_i$.

\noindent\textbf{Effect of entropy-based adapter selection:} Entropy serves as a natural indicator of adapter-task alignment, when an adapter properly matches the input task, it generates confident, peaked predictions (low entropy), whereas mismatched adapters produce uncertain, flat distributions (high entropy). This intrinsic property makes entropy a reliable metric for selecting the most suitable adapter.

\subsection{Task-Specific and Universal Model Ensemble}  While task-specific adapters excel at extracting discriminative features for individual tasks, their narrow focus often fails to capture transferable patterns that could aid in distinguishing visually similar classes across different tasks. Our objective is to leverage both specialized and general features effectively, enabling better discrimination between visually similar classes from distinct tasks. Building on this insight, we propose a novel inference strategy: given a test image, we not only select the most suitable task-specific adapter 
$\mathcal{A}^*$ according to Eq.~\ref{eq:min_entropy} but also incorporate the predictions generated by the universal adapter to enhance classification robustness:
\begin{align} \label{eq:prediction}
     y^* = \argmax_y \left( f_y(\mathbf{x};\mathcal{A}^*) + f_y(\mathbf{x};\mathcal{A}_{\text{uni}}) \right).
\end{align}

\begin{table*}[t]
	\caption{ Average and last performance comparison on four datasets with {\bf ViT-B/16-IN21K} as the backbone. 
		We report all compared methods with their source code.
		The best performance is highlighted in bold. None of the methods utilize exemplars in their implementation.
	}\label{tab:benchmark}
	\centering
	\small 
	\begin{tabular}{@{}lc  cccccccccc cccccccc}
		\toprule
		\multicolumn{1}{l}{\multirow{2}{*}{Method}} & 
		
		\multicolumn{2}{c}{CIFAR  B0 Inc5} 
		& \multicolumn{2}{c}{ImageNet-R B0 Inc20}
		& \multicolumn{2}{c}{ImageNet-A B0 Inc20}
		& \multicolumn{2}{c}{ObjectNet B0 Inc20}
		\\
		 & {$\bar{\mathcal{A}}$} & ${\mathcal{A}_B}$  
        & {$\bar{\mathcal{A}}$} & ${\mathcal{A}_B}$   
        & {$\bar{\mathcal{A}}$} & ${\mathcal{A}_B}$ 
        & {$\bar{\mathcal{A}}$} & ${\mathcal{A}_B}$ \\
		
		\midrule
		L2P~\cite{wang2022learning}    & 85.94 & 79.93 & 75.46 & 69.77 &  49.39 & 41.71 &  63.78 & 52.19 \\
		DualPrompt~\cite{wang2022dualprompt}    &  87.87 & 81.15 &73.10 &67.18  & 53.71 & 41.67 & 59.27 & 49.33 \\
		CODA-Prompt~\cite{Smith_2023_CVPR}  & 89.11 & 81.96 & 77.97 &72.27 & 53.54 & 42.73 & 66.07 &53.29 \\
		SLCA~\cite{zhang2023slca}  & 92.49  & 88.55   &81.17  &77.00  &68.66  & 58.74& 72.55  & 61.30 \\ 
        SSIAT~\cite{Tan_2024_CVPR}  & 93.52  & 90.07   &83.20 &78.85  &70.83  & 62.23& 73.65  & 62.45 \\
        MOS~\cite{sun2024mos}  & 93.30  & 89.25   & 82.96 &77.93  & 67.08 & 56.22 &74.69	& 63.62  \\
		
		SimpleCIL~\cite{zhou2024revisiting}   &   87.57 & 81.26 & 61.26 &54.55  & 59.77 & 48.91 & 65.45 & 53.59 \\
		APER + Adapter~\cite{zhou2024revisiting} &   90.65 &  85.15&75.82  &67.95 & 60.47 &49.37 &  67.18 & 55.24 \\
		RanPAC~\cite{McDonnell_2023_NeurIPS} &  94.00  & 90.62   &82.98  &77.94 &69.32  & 61.82& 72.76  & 62.02 \\
		EASE~\cite{zhou2024expandable}  & 91.51 &85.80  & 81.74 & 76.17 & 65.34 & 55.04 & 70.84 & 57.86 \\
		
		\midrule
		TUNA (Ours)  & \bf94.44  & \bf90.74   & \bf84.22 &\bf79.42  & \bf73.78 & \bf64.78 &\bf76.46	& \bf66.32  \\
		\bottomrule
	\end{tabular}
\end{table*}
  \begin{figure*}[t]
	\centering
	\begin{subfigure}{0.23\linewidth} 
		\includegraphics[width=\linewidth]{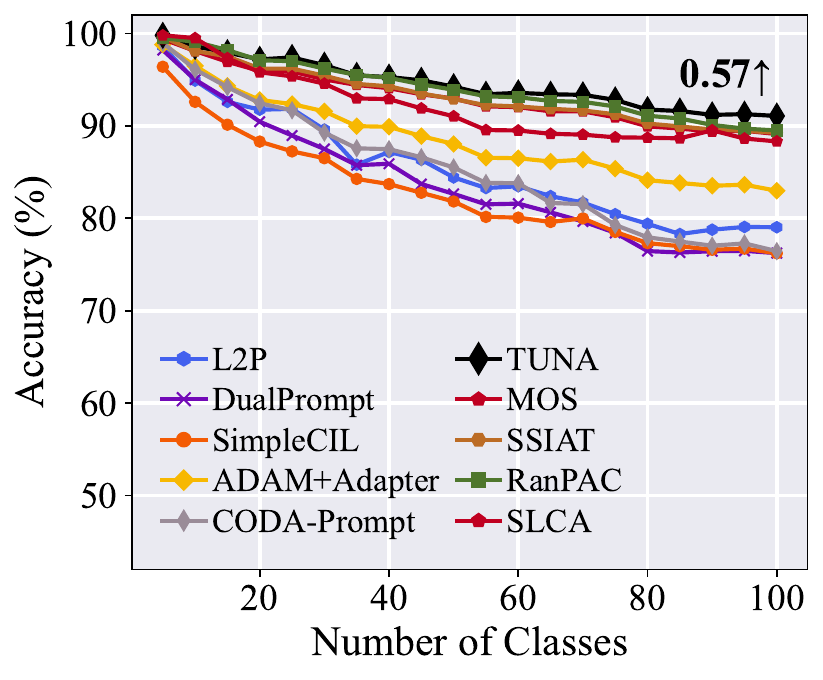}
		\caption{\small CIFAR B0 Inc5}
		\label{fig:benchmark-cifar}
	\end{subfigure}
	\hfill
	\begin{subfigure}{0.23\linewidth}
		\includegraphics[width=\linewidth]{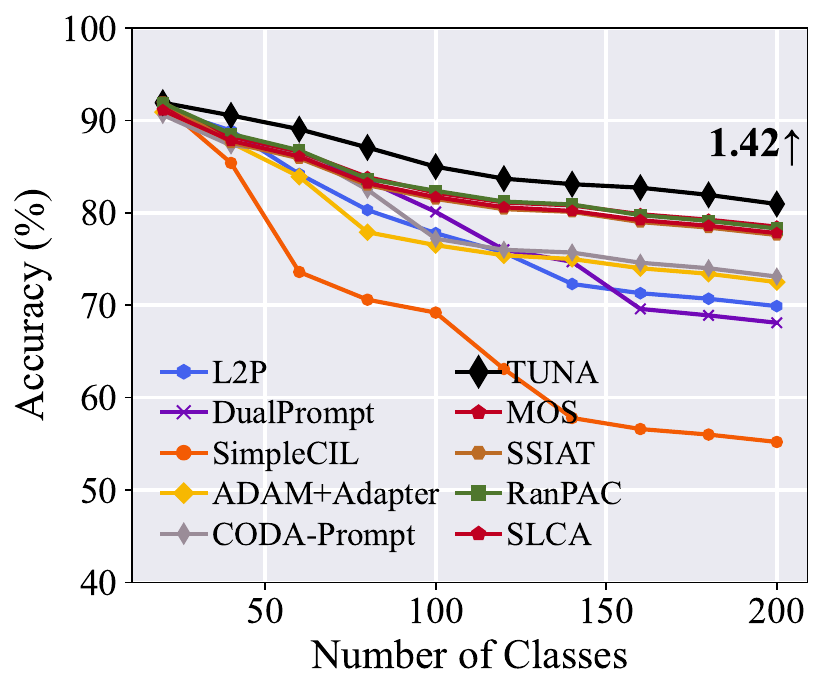}
		\caption{\small ImageNet-R B0 Inc20}
		\label{fig:benchmark-inr}
	\end{subfigure}
	\hfill
	\begin{subfigure}{0.23\linewidth}
		\includegraphics[width=\linewidth]{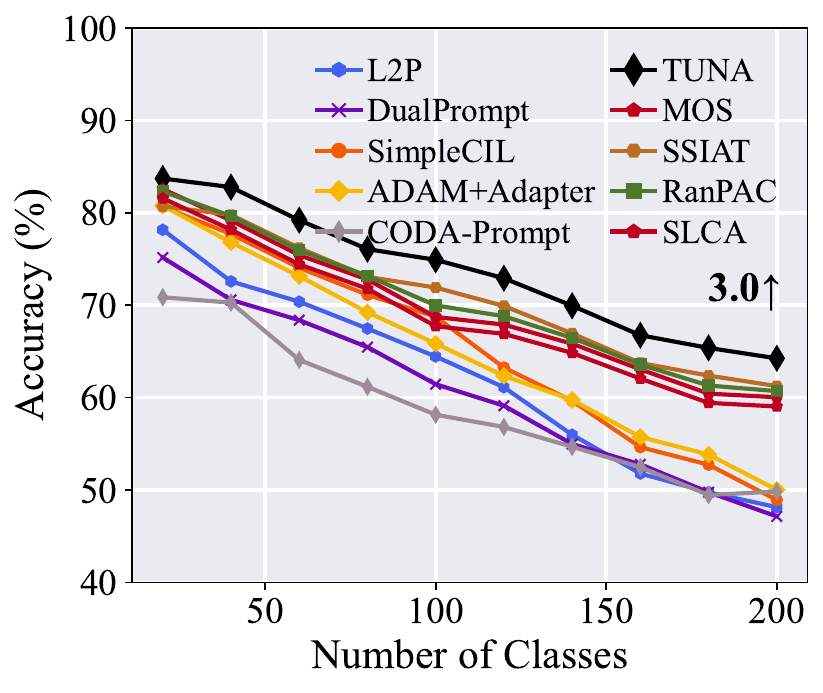}
		\caption{\small ImageNet-A B0 Inc20}
		\label{fig:benchmark-ina}
	\end{subfigure}
	\hfill
	\begin{subfigure}{0.23\linewidth}
		\includegraphics[width=\linewidth]{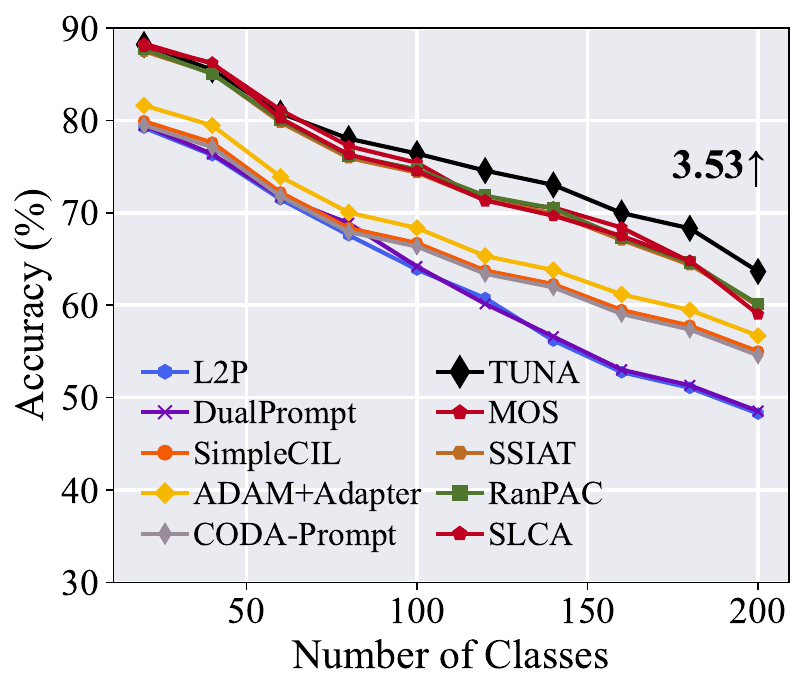}
		\caption{\small ObjectNet B0 Inc20}
		\label{fig:benchmark-obj}
	\end{subfigure}
	
	\caption{\small Performance curve of different methods under different settings. All methods are initialized with {\bf ViT-B/16-IN1K}. The relative improvement over the second-best method is annotated with numerical values above the curves at the final incremental stage. }
	\label{fig:benchmark}
 \vspace{-3mm}
\end{figure*}

\noindent\textbf{Summary of TUNA:} As illustrated in Figure~\ref{figure:method}, we initialize and train an adapter for each incremental task to encode the task-specific information, and then we compute the class-wise mean and variance upon completing the training of each task-specific adapter. These statistical features are subsequently replayed during future incremental learning tasks to alleviate catastrophic forgetting in the classification head. Finally, we fuse these task-specific adapters into a universal adapter, which amalgamates cross-task knowledge while preserving domain-invariant representations. During the inference phase, we employs an entropy-guided adapter selection mechanism that combines the most confident task-specific adapter with the universal adapter to generate more accurate predictions.

\section{Experiments}

In this section, we conduct a thorough evaluation of our proposed method using four benchmark datasets, comparing its performance against state-of-the-art methods to demonstrate its advantages. Additionally, we provide an ablation study and further analysis to validate the robustness and effectiveness of our approach.

\subsection{Implementation Details}
\noindent\textbf{Dataset:} Given that pre-trained models  encapsulate extensive knowledge from upstream tasks, we adopt the evaluation framework proposed in \cite{zhou2024revisiting} to assess the performance on various benchmark datasets, including CIFAR100~\cite{krizhevsky2009learning},  ImageNet-R~\cite{hendrycks2021many}, ImageNet-A~\cite{hendrycks2021unnatural}, 
 and ObjectNet~\cite{barbu2019objectnet}. These datasets represent typical CIL benchmarks and include out-of-distribution datasets that exhibit a \emph{significant domain gap} relative to ImageNet. Specifically, there are 100 classes in CIFAR100, 200 classes in ImageNet-R, ImageNet-A and ObjectNet. 
 
\noindent\textbf{Dataset split:} In accordance with the benchmark protocols established in~\cite{Rebuffi_2017_CVPR}, we employ the notation `B-$m$ Inc-$n$' to represent class splits, where $m$ indicates the number of classes in the initial task, and $n$ denotes the number of classes in each subsequent incremental task.  To ensure a fair and consistent comparison, we follow~\cite{Rebuffi_2017_CVPR} and randomly shuffle class orders using a random seed of 1993 before splitting the data. 
We ensure consistency in the training and testing sets across all methods, following ~\cite{zhou2024revisiting,Tan_2024_CVPR,zhou2024expandable}..

\noindent\textbf{Comparison methods:} We compare our approach with state-of-the-art PTM-based CIL methods, including prompt-based techniques (L2P~\cite{wang2022learning}, DualPrompt~\cite{wang2022dualprompt}, and CODA-Prompt~\cite{Smith_2023_CVPR}), full-model fine-tuning approaches like SLCA~\cite{zhang2023slca}, and adapter-based methods such as SSIAT~\cite{Tan_2024_CVPR}, EASE~\cite{zhou2024expandable}, and MOS~\cite{sun2024mos}. We also consider prototype-based SimpleCIL~\cite{zhou2024revisiting} and first-session adaptation approaches including RanPAC~\cite{McDonnell_2023_NeurIPS} and APER~\cite{zhou2024revisiting}. All comparative methods employ identical pre-trained models and experimental setups to guarantee fair comparison.

\noindent\textbf{Training details:} We use PyTorch~\cite{paszke2019pytorch} to implement all models on NVIDIA RTX 4090 with the \emph{same} network backbone. Since the wide range of PTMs are publicly accessible~\cite{rwightman2020timm}, we choose two representative models following~\cite{zhou2024revisiting}, denoted as \textbf{ViT-B/16-IN1K} and \textbf{ViT-B/16-IN21K}. They are both initially pre-trained on ImageNet21K, while the former is further finetuned on ImageNet1K. In our method, we set the batch size to 48 and train for 20 epochs using the SGD optimizer with momentum. The learning rate is initially set to 0.01 and follows a cosine annealing decay pattern. The projection dimension $r$ in the adapter is set to 16, the weight $\lambda$ in Eq.~\ref{eq:adapter-overall} is initialized at 1e-3 and follows an exponential decay schedule.

\noindent\textbf{Evaluation protocol:} Following the benchmark established by~\cite{Rebuffi_2017_CVPR}, we denote the Top-1 accuracy after the $b$-th stage as $\mathcal{A}_b$. Moreover, we use $\mathcal{A}_B$ (the performance after the last stage) and $\bar{\mathcal{A}}=\frac{1}{B}\sum_{b=1}^{B}\mathcal{A}_b$ (average performance along incremental stages) as measurements.

\subsection{Benchmark Comparison}
 In this section, we conduct a comprehensive comparison of our proposed method, TUNA, against state-of-the-art approaches on four benchmark datasets and different backbone weights. Table~\ref{tab:benchmark} reports the comparison of different methods with ViT-B/16-IN21K. We can infer that our method achieves the best performance among all four benchmarks,
 substantially outperforming the current SOTA methods. We also report the incremental performance trend of different methods in Figure~\ref{fig:benchmark} with ViT-B/16-IN1K. As annotated at the end of each image, we find our method consistently outperforms the runner-up method, further underscoring its effectiveness. 

\begin{figure}[t]
	\centering
	\begin{subfigure}{0.49\linewidth}
		\includegraphics[width=1\columnwidth]{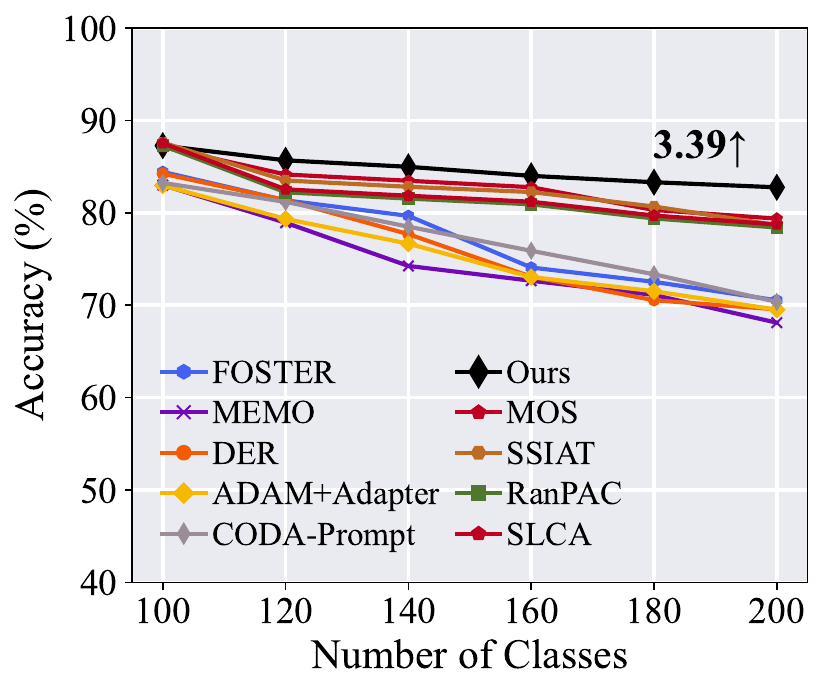}
		\caption{\small ImageNet-R B100 Inc20}
		\label{fig:benchmark-inrb100inc50}
	\end{subfigure}
	\hfill
	\begin{subfigure}{0.49\linewidth}
		\includegraphics[width=1\linewidth]{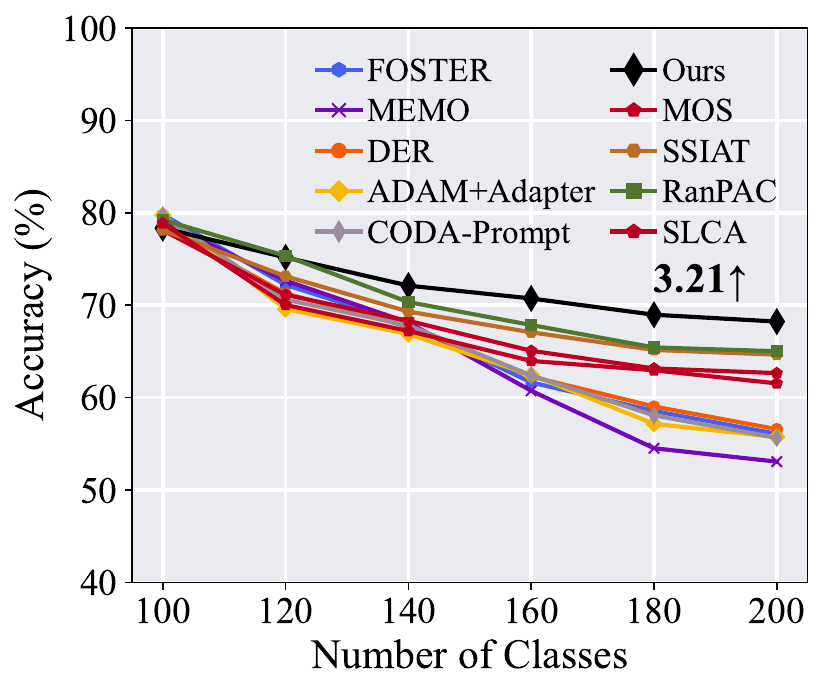}
		\caption{\small ImageNet-A B100 Inc20}
		\label{figure:benchmark-inrb100inc50}
	\end{subfigure}
	\caption{\small  Experimental results on ImageNet-R and ImageNet-A with large base classes. All methods are based on the same PTM.} 
	\label{fig:benchmark-large-base}
 \vspace{-3mm}
\end{figure}

 To further validate the robustness of our approach, we extend our evaluation beyond the standard B0 benchmark (presented in Table~\ref{tab:benchmark} and Figure~\ref{fig:benchmark}) to a  large-base setting. In Figure~\ref{fig:benchmark-large-base}, we compare our method with several SOTA methods with vast base classes. As shown
 in Figure~\ref{fig:benchmark-large-base}, TUNA still outperforms other methods. Additionally, we also compare TUNA to traditional CIL methods such as iCaRL~\cite{Rebuffi_2017_CVPR}, 
 DER~\cite{yan2021dynamically}, FOSTER~\cite{wang2022foster}, MEMO~\cite{zhou2022model}, TagFex~\cite{zheng2025task} by implementing them with the same pre-trained weight in Table~\ref{tab:benchmark-typicalmethods}. Notably, TUNA maintains its leading performance, achieving a higher average accuracy than the closest competitor while remaining exemplar-free—a key advantage in memory-constrained scenarios.

 \begin{table}[t]
	\vspace{-3mm}
	\caption{ In contrast to conventional exemplar-based continual learning approaches, TUNA operates without storing any exemplars. All compared methods utilize the identical pre-trained backbone architecture (ViT-B/16-IN21K) for fair evaluation.
	}  
	\label{tab:benchmark-typicalmethods}
	\centering
	\vspace{-3mm}
	\resizebox{0.98\columnwidth}{!}{%
		\begin{tabular}{@{}lccccccccc }
			\toprule
			\multicolumn{1}{l}{\multirow{2}{*}{Method}} &
			\multicolumn{1}{l}{\multirow{2}{*}{Exemplars}} & 
			\multicolumn{2}{c}{ImageNet-R B0 Inc20} & \multicolumn{2}{c}{CIFAR B0 Inc10}  \\
			& & {$\bar{\mathcal{A}}$} & ${\mathcal{A}_B}$  
			& {$\bar{\mathcal{A}}$} & ${\mathcal{A}_B}$
			\\
			\midrule
			iCaRL~\cite{Rebuffi_2017_CVPR}& 20 / class & 72.42&60.67 & 82.46& 73.87 \\
			DER~\cite{yan2021dynamically} &20 / class  & 80.48& 74.32 & 86.04& 77.93\\
			FOSTER~\cite{wang2022foster} &20 / class   & 81.34&74.48 &89.87& 84.91\\
			MEMO~\cite{zhou2022model} &20 / class &74.80& 66.62 & 84.08& 75.79\\
            TagFex~\cite{zheng2025task} &20 / class &83.23& 78.45 & 92.17& 89.26\\
			\midrule
			TUNA  &\bf 0 & \bf 85.90& \bf 80.95  & \bf 95.05 & \bf 92.15\\
			\bottomrule
		\end{tabular}
	}
	\vspace{-3mm}
\end{table}
  It is important to highlight that traditional CIL methods rely on storing exemplars to retain previously learned knowledge, whereas our approach eliminates this requirement. We follow~\cite{Rebuffi_2017_CVPR} to set the exemplar number to 20 per class for these methods. TUNA still works competitively
 in comparison to these exemplar-based methods.
 
 \begin{figure}[t]
	\centering
	\begin{subfigure}{0.49\linewidth}
		\includegraphics[width=1\columnwidth]{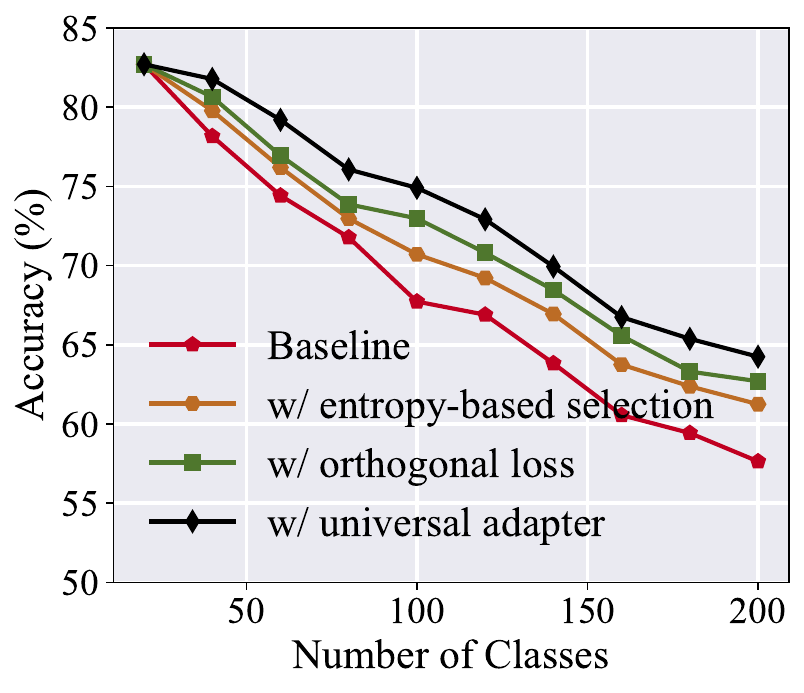}
		\caption{\small Ablation study}
		\label{fig:ablation}
	\end{subfigure}
	\hfill
	\begin{subfigure}{0.49\linewidth}
		\includegraphics[width=1\linewidth]{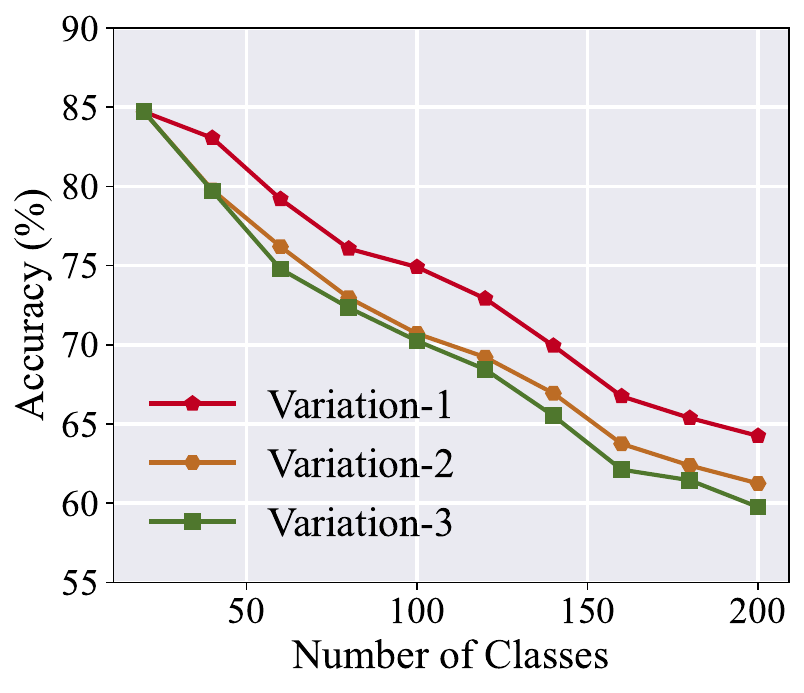}
		\caption{\small Inference ablation}
		\label{figure:inference}
	\end{subfigure}
	\caption{\small \textbf{Left:} Ablation study of different components in TUNA. We find each component contributes to enhancing the performance. \textbf{Right:} Experimental results on ImageNet-A B0 inc20 with different inference strategies.} 
	\label{fig:inference}
 \vspace{-3mm}
\end{figure}
\subsection{Ablation Study}
In this section, we perform an ablation study to evaluate the contribution of each component in TUNA. Specifically, we present the incremental performance of various configurations on ImageNet-A B0 Inc20 in Figure~\ref{fig:ablation}. In the figure, ‘Baseline' denotes training task-specific adapters for each task and predicting using all adapters during inference, selecting the maximum logit as the final prediction. ‘w/ entropy-based adapter selection' means selecting the task-specific adapter based on entropy and using its output for prediction, which proves to be an effective strategy for choosing the appropriate adapter. Furthermore, ‘w/ orth loss' introduces an orthogonality loss during training to enhance task-specific knowledge learning, and the results show that this addition improves performance. Finally, ‘w/ universal adapter' ensembles the outputs from a universal adapter, which captures general knowledge shared across tasks, enabling the model to better handle all encountered tasks. The ablation study confirms that each component in TUNA contributes to improving CIL performance.
\subsection{Further Analysis}
\noindent\textbf{Different inference strategies}:
To validate our proposed inference strategy, we conduct experiments on ImageNet-A B0 Inc20 using three inference strategies: Variation-1 (our strategy), Variation-2 (task-specific adapter selection based on entropy), and Variation-3 (sole reliance on the universal adapter). As shown in Figure~\ref{figure:inference}, Variation-1 consistently outperforms the others across all tasks. Variation-2 fails to leverage shared knowledge between tasks, while Variation-3 lacks the granularity to capture task-specific nuances, resulting in suboptimal performance.

\begin{figure}[t]
    \centering
    \begin{subfigure}[t]{0.49\linewidth}
        \vspace{-5pt} 
        \includegraphics[width=\linewidth]{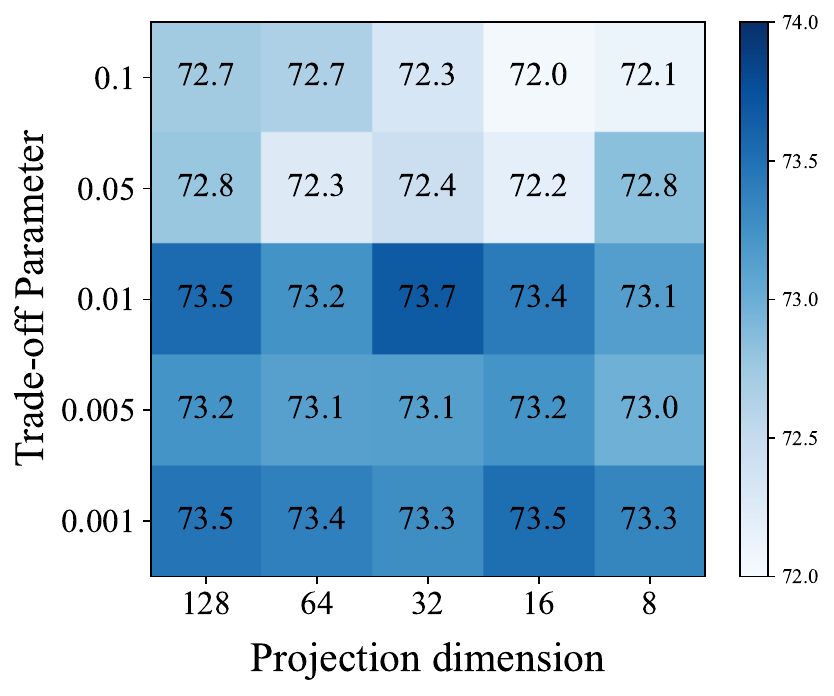}
        \vspace{-0.5\baselineskip} 
        \caption{\small Hyperparameters robustness}
        \label{fig:Robustness of hyperparameters}
    \end{subfigure}
    \hfill
    \begin{subfigure}[t]{0.49\linewidth}
        \vspace{-5pt} 
        \includegraphics[width=\linewidth]{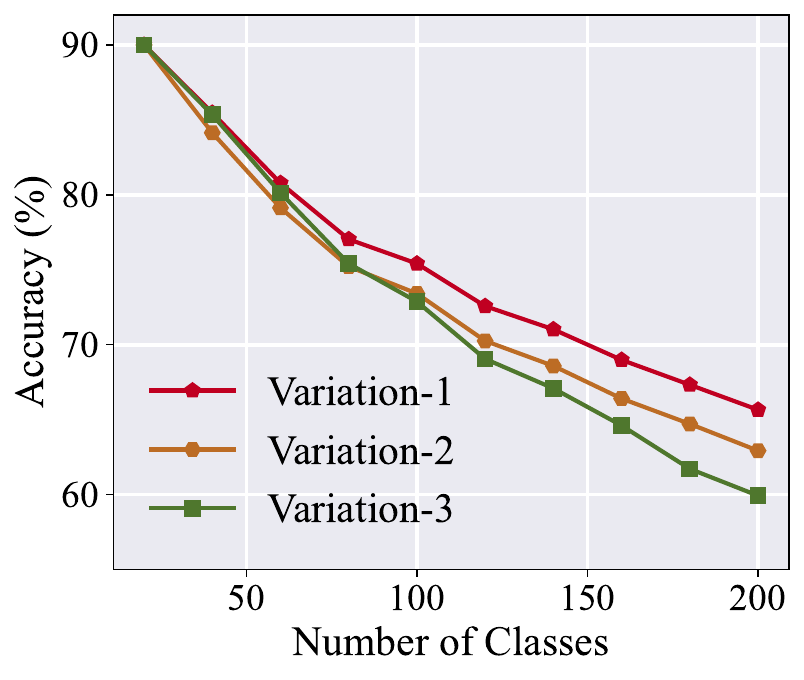}
        \vspace{-0.6\baselineskip} 
        \caption{\small Variations of Eq.~\ref{eq:adapter-orth}}
        \label{figure:orth}
    \end{subfigure}
    \caption{\small Further analysis on parameter robustness and orthogonal loss implementation.} 
    \label{fig:further analysis}
    \vspace{-3mm}
\end{figure}

\noindent\textbf{Parameter robustness}: TUNA involves two hyperparameters, the projection dim $r$ in the adapter and the trade-off parameter $\lambda$ in Eq.~\ref{eq:adapter-overall}. To evaluate their robustness, we conduct experiments on ImageNet-A B0 Inc20 by varying these parameters. Specifically, we choose $r$ among $\{8, 16, 32, 64, 128\}$, and $\lambda$ among $\{0.001, 0.005, 0.01, 0.05, 0.1\}$. We report the average performance in Figure~\ref{fig:Robustness of hyperparameters}. The results demonstrate that the performance remains stable across different parameter values.

\noindent\textbf{Different orthogonal loss}: In Eq.~\ref{eq:adapter-orth}, we force the current adapter's up projection weight to be orthogonal to previous adapters' up projection weights, we call it Variation-1. Additionally, we can extend this orthogonality constraint to the current adapter's down projection weight relative to previous adapters' down projection weights, termed Variation-2, or apply it to both the up projection and down projection weights simultaneously, denoted as Variation-3. We conduct experiments on ObjectNet B0 Inc20 setting to compare different losses. As we can see from Figure~\ref{figure:orth}, with other settings the same, we find Variation-1 performs the best among these variations. This is likely because the up projection weight plays a more critical role in capturing task-specific features, and enforcing orthogonality on it alone is sufficient to reduce task interference. In contrast, the down projection weight primarily projects input features into a lower-dimensional space, and overly restricting it may hinder the model's ability to encode task-specific information. Additionally, applying constraints to both weights simultaneously may introduce excessive rigidity, reducing flexibility and risking underfitting. Thus, focusing orthogonality constraints solely on the up projection weight offers a more balanced and efficient approach for continual learning.

\begin{figure}
\vspace{-5mm}
	\centering
	\includegraphics[width=1.0\columnwidth]{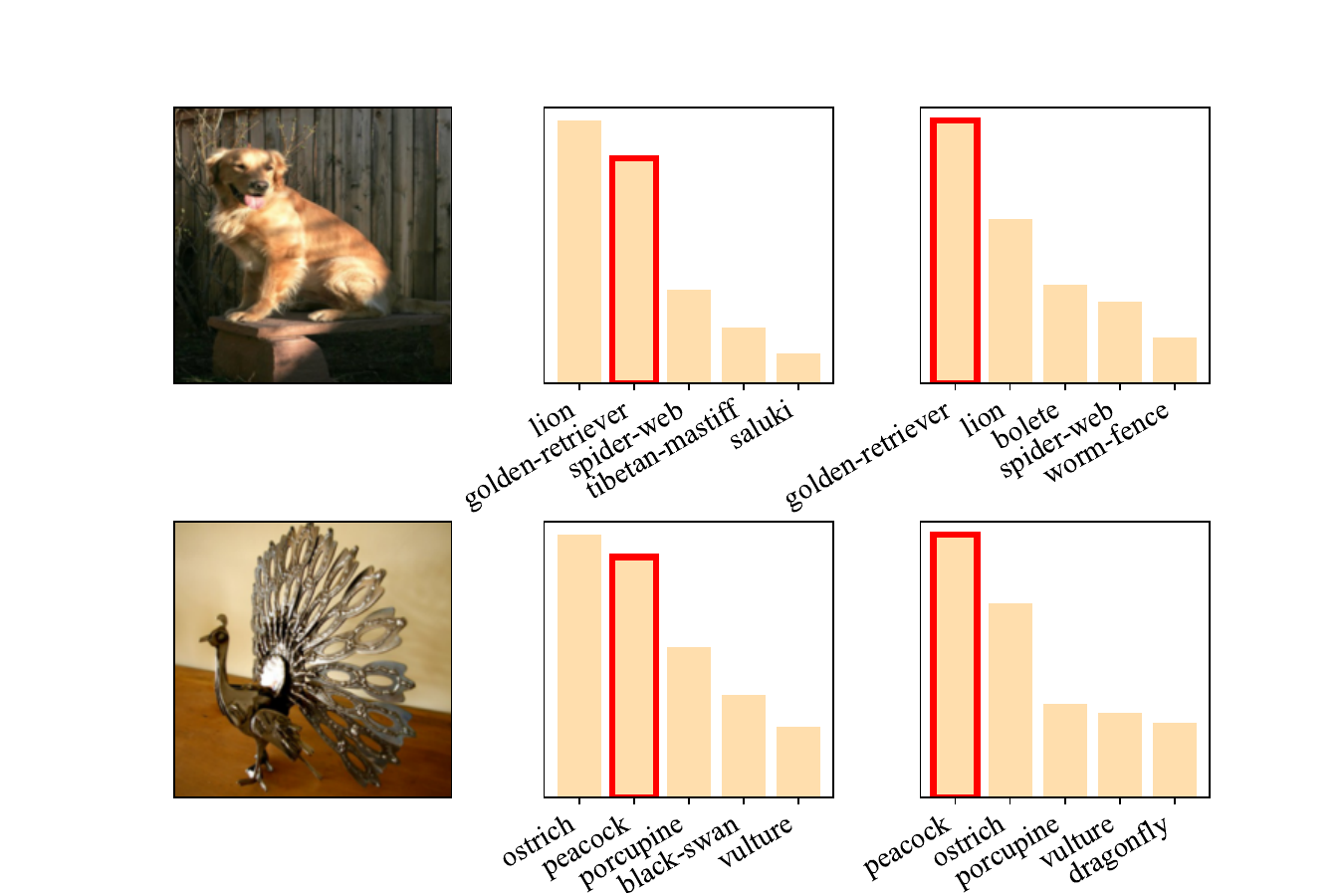}
	\caption{ Visualizations of the predictions on ImageNet-R. The original images are depicted in the first column, followed by the top-5 prediction probability produced by task-specific adapter, and the probabilities generated by the universal adapter in the last column. The ground-truth class is highlighted with red boxes.}
	\label{fig:diff}
 \vspace{-3mm}
\end{figure}

\noindent\textbf{Visualizations}: To explore why combining task-specific and universal adapters boosts performance, we visualize predictions from each adapter separately using ImageNet-R images and the model trained under the B0 Inc20 setting. Figure~\ref{fig:diff} shows the original images in the first column, top-5 predictions from task-specific adapters in the second column, and top-5 predictions from the universal adapter in the third column. Task-specific adapters, focusing on limited information, often misclassify similar classes, such as predicting a golden retriever as a lion or  a peacock as an ostrich. In contrast, the universal adapter, which integrates cross-task knowledge, captures shared features and refines predictions, increasing the chances of correct classification. This synergy enhances overall performance.

\section{Conclusion}
Incremental learning is crucial for practical systems. This paper introduces a novel method that integrates Task-Specific and Universal Adapters(TUNA) for pre-trained model-based CIL. Specifically, we train task-specific adapters to capture distinct features for their tasks. We also introduce an adapter fusion mechanism to create a universal adapter that encapsulates shared knowledge across tasks. During inference, we employ an entropy-based selection to choose the most suitable task-specific adapter and then ensemble its predictions with those from the universal adapter. Extensive
 experiments verify TUNA’s effectiveness.

\noindent\textbf{Limitations and future works:}  The process of selecting the optimal task-specific adapter requires multiple forward passes through the model, resulting in increased computational time.  Future works
 include designing methods to speed up the algorithm.

\section*{Acknowledgments}

This work is supported by the NSFC (62376118), CCF-Tencent Rhino-Bird Open Research Fund (RAGR20240101), Fundamental Research Funds for the Central Universities (14380021), and the Collaborative Innovation Center of Novel Software Technology and Industrialization. 

{
    \small
    \bibliographystyle{ieeenat_fullname}
   \bibliography{main}
}

\end{document}